\documentclass{article} 
\pdfoutput=1 
\providecommand{\keywords}[1]{\textbf{Keywords:} #1}

\usepackage{graphicx} % more modern
\usepackage{subfigure} 
\usepackage{fullpage}

\usepackage{algpseudocode}
\usepackage{algorithm}
\usepackage{subfigure}
\usepackage{color}
\usepackage{url}

\usepackage{amsmath,amssymb,amsthm,bm}
\DeclareMathOperator*{\argmax}{argmax}

\usepackage{accents}

\title{GLAD: Group Anomaly Detection in Social Media Analysis- \\ Extended Abstract \\ }
\date{ }

\author{
Qi(Rose) Yu \\
Department of Computer Science \\
University of Southern California \\
\texttt{qiyu@usc.edu}
\and
Xinran He \\
Department of Computer Science \\
University of Southern California \\
\texttt{xinranhe@usc.edu}
\and
Yan Liu \\
Department of Computer Science \\
University of Southern California \\
\texttt{yanliu.cs@usc.edu}
}

\begin{document} 
\maketitle
\maketitle

\begin{abstract}
Traditional anomaly detection on social media mostly focuses on individual point anomalies while anomalous phenomena usually occur in groups.  Therefore it is valuable to study the collective behavior of individuals and detect group anomalies.  Existing group anomaly detection approaches rely on the assumption that the groups are known, which can hardly be true in real world social media applications. In this paper, we take a generative approach by proposing a hierarchical Bayes model: \textit{Group Latent Anomaly Detection} (GLAD) model. GLAD takes both pair-wise and point-wise data as input, automatically infers the groups and detects group anomalies simultaneously. To account for the dynamic properties of the social media data, we further generalize GLAD to its dynamic extension d-GLAD.  We conduct extensive experiments to evaluate our models on both synthetic and real world datasets. The empirical results demonstrate that our approach is effective and robust in discovering latent groups and detecting group anomalies.
\end{abstract}

\section{Introduction}

Social media provide convenient platforms for people to share, communicate, and collaborate. While people enjoy the openness and convenience of social media, many malicious behaviors, such as bullying, terrorist attack planning, and fraud information dissemination, can happen. Therefore, it is extremely important that we can detect these abnormal activities as \textit{accurately} and \textit{early} as possible to prevent disasters and attacks.

By definition, anomaly detection aims to find ``an observation that deviates so much from other observations as to arouse suspicion that it was generated by a different mechanism'' \cite{hawkins_identification_1980}. Several algorithms have been developed specifically for social media anomaly detection such as power-law models \cite{akoglu_anomaly_2009}, spectral decomposition \cite{von2007tutorial}, scan statistics \cite{priebe_scan_2005}, and random walk \cite{pan_automatic_2004,tong2008random}. However, these algorithms only detect the individual \textit{ point anomaly}. For example, \cite{akoglu_anomaly_2009} proposes an ``OddBall'' algorithm to spot anomalous nodes in a graph. The algorithm extracts features from the egonet of the node and declares anomaly node whose features deviate from the power-law pattern.

In reality, anomaly may not only appear as an individual point, but also as a group. For instance, a group of people collude to create false
product reviews or threat campaign in social media platforms; in large organizations, malfunctioning teams or insider groups closely coordinate with each other to achieve a malicious goal. Those appear as examples for another type of anomaly: \textit{group anomaly}, which has not been thoroughly examined in social media analysis. In this work, we focus on group anomaly detection. We are interested in finding the groups which exhibit a pattern that does not conform to the majority of other groups. This problem has found its applications in galaxy identification \cite{xiong_hierarchical_2011}, high energy particle physics \cite{muandet2013one}, anomalous image detection and turbulence vorticity modeling \cite{xiong_group_2011}.

We identify three major challenges in group anomaly detection: (\romannumeral 1)
Two forms of data coexist in social media: one is the point-wise data, which characterize the features of an individual person. The other is pair-wise relational data, which describe the properties of social ties. In social science, a fundamental axiom of social media analysis is the concept that structure matters. For example, teams with the same composition of member skills can perform very differently depending on the patterns of relationships among the members \cite{ borgatti_network_2009}. Therefore, it is important to take into account both point-wise and pair-wise data during anomaly detection.
(\romannumeral 2) Group anomaly is usually more subtle than individual anomaly. At the individual level, the activities might appear to be normal \cite{chandola2009anomaly}. Therefore, existing anomaly detection algorithms usually fail when the anomaly is related to a group rather than individuals.
(\romannumeral 3)
Empirical studies in social media analysis suggest the dynamic nature of individual network positions \cite{kossinets_empirical_2006}. People's activities and communications change constantly over time and we can hardly know the groups beforehand. Thus developing a method that can be easily generalized to dynamic setting is critical to anomaly detection in evolving social media data.
%understanding the evolution of social networks.

In this paper, we take a  graphical model approach to address those challenges. We propose a hierarchical model, i.e, \textit{Group Latent Anomaly Detection} (GLAD) model, to connect two forms of data. To handle the dynamic characteristics of the social media data, we further develop a dynamic extension of GLAD: the d-GLAD model. We show that GLAD outperforms existing approaches in terms of group anomaly detection accuracy and robustness. When dealing with dynamic social networks, the dynamic extension of GLAD achieves lower false positive rate and better data fitting. %relying on the assumption that the changes will not largely affect the overall constitution of the groups.   %To do so, we present detailed inference and learning procedures for our models and perform extensive empirical valuation on several datasets.
The major contributions of this paper can be summarized as follows:
\begin{enumerate}
\item We formulate the problem of group anomaly detection in the context of social media analysis for both static and dynamic settings and articulate the three major challenges associated with the task.

\item We develop a graphical model called GLAD. GLAD can successfully discover the group structure of social media and detect group anomalies. We also generalize GLAD to its dynamic extension and provide tractable model inference algorithms.

\item We conduct thorough experiments on both synthetic and real world datasets using anomaly injections. We also construct a meaningful dataset from ACM publication dataset for rigorous evaluation. The dataset is accessible at \url{http://www-bcf.usc.edu/~liu32/data.html}.  % Due to the lack of group anomaly, we conduct anomaly injection to evaluate the detection performance. The results from US congress senator voting data set and scientific publication datasets justify the advantages of our approach in terms of anomaly detection, group recovery and data fitting tasks. Furthermore, we share with the community a well-processed ACM publication dataset with dynamic network structure and bag of word features.
\end{enumerate}

\section{Related Work}

We review the related models on group anomaly detection and illustrate the motivation behind our approach.

The Multinomial Genre Model (MGM)  proposed in  \cite{xiong_hierarchical_2011} first investigates the problem following the paradigm of Latent Dirichlet Allocation (LDA) \cite{blei_latent_2003}. As a text processing tool, LDA assumes that each word is associated with a topic and a document is a mixture of topics. Similarly, MGM models a group as a mixture of Gaussian distributed topics with certain mixture rate and assumes there exists ``best" mixture rates, corresponding to the mixture rates of normal groups. Then it conducts group anomaly detection by scoring the mixture rate likelihood of each group. One drawback of MGM is that the set of candidate mixture rates is shared globally by groups. It might leads to poor performance when groups have different sets of mixture rates. \cite{xiong_group_2011} further extends MGM to Flexible Genre Model (FGM)  with more flexibility in the generation of topics. Specifically, the model considers the set of topic mixture rates as random variables rather than model hyper-parameters, which would adapt to diverse ``genres'' in groups, each of which is a typical distribution of topic mixture rates.

Another line of work takes a discriminative approach. \cite{muandet2013one} uses the same definition of group anomaly from \cite{xiong_hierarchical_2011}. It considers kernel embedding of the probabilistic distributions and generalizes one-class support vector machine from point anomaly detection to group anomaly detection. The proposed support measure machine (SMM) algorithm maps the distributions to a probability measure space with kernel methods, which can handle the aggregate behavior of data points.

However, existing approaches separate the group anomaly detection task into two stages: group discovery and anomaly detection. They require the group information to be given before applying the anomaly detection algorithms. For example, in  \cite{xiong_hierarchical_2011}, the Sloan Digital Sky Survey (SDSS) dataset needs to be pre-processed before feeding into MGM. The authors first construct a neighborhood graph and then treat the connected components in the graph as groups. For the application on turbulence data, the FGM model \cite{xiong_group_2011} considers the vertices in a local cubic region as a group. In SMM \cite{muandet2013one}, the authors treat the high energy particles generated from the same collision event as a group.

The two-stage approaches identify the groups from the pair-wise data and infer the anomalies based on the point-wise data. This strategy assumes that the point-wise and pair-wise data are marginally independent. However, such independence assumption might underestimate the mutual influence between the group structure and the  feature attributes. The detected group anomalies can hardly reveal the joint effect of these two forms of data. These motivate us to build an alla prima that can account for both forms of data and accomplish the tasks of group discovery and anomaly detection all at once.

Additionally, existing work can only deal with static network and fixed size groups. This is not feasible for the time-evolving nature of social media data. For example, in corporate networks, employees may switch teams from one to the other. The organization structure of a team may also change. As the dynamic setting needs to take into account the flexible group size and the changing mixture rates, we further adapt our model to the dynamic setting and formulate the problem as a change point detection task. 

Group anomaly detection in social media analysis may shed light on a wide range of real world problems such as corporate restructuring, team job-hopping and political inclination shift to which our approach can apply. In section \ref{sec:definition}, we provide a formal definition of group anomaly in social media analysis. We first develop GLAD$^0$ as well as its learning and inference algorithm in section \ref{sec:glad0}. Then we present a computationally more efficient model design: GLAD in section \ref{sec:glad}. In section \ref{sec:dglad}, we describe the dynamic GLAD model: d-GLAD, which can handle the dynamic social networks. Section \ref{sec:experiment} shows the empirical evaluation results of GLAD and d-GLAD on synthetic and real world datasets compared with existing baseline models.

\section{Definition of Group Anomaly}
\label{sec:definition}
% definition of group anomaly
The core of our group anomaly definition lies in the collective behavior of individuals. For example, a document is a mixture of various topics and a team is a mixture of different roles. Therefore, we model the node features of each group as a mixture of components. Each component could be an article topic, a social role or a job title. Specifically, we can describe a component as either a discrete variable such as multinomial distribution or a continues variable like Gaussian  distribution, depending on the data type of features.  Here we use the term \textit{role} as a general notion for the component. We assume that there are a fixed number of roles and each of which denotes a particular distribution of node features. All groups share the same set of roles but possibly with different role mixture rates. Normal groups follow the same pattern with respect to their role mixture rates, but the anomalous group has a role mixture rate that deviates from the normal pattern. 

For the static GLAD model, we are interested in the distribution of the role mixture rates across the groups. According to our assumption, the mixture rates of normal groups are more likely to appear. For groups with very rare role mixture rates, we treat them as group anomalies. One example of this type of group anomaly comes from particle physics. It is widely accepted that the dynamics of known particles are governed by the \textit{Standard Model}, which corresponds to the normal pattern. Unknown particles would contaminate the distribution of the Standard Model. Detecting those anomalies could potentially lead to the discovery of new physical phenomenon. In practice, we first identify the normal mixture rates. Then for each learned group, we evaluate the likelihood of its observations being generated with the normal mixture rates. The lower the likelihood value is, the more anomalous the group would be.

For the dynamic d-GLAD model, we emphasize on the temporal aspect of the data and detect the change of the role mixture rate within the groups. For instance, in scientific area, it is valuable to study the evolution of research topics and detect the bursty time periods. In the dynamic setting, since the structure of groups change as well as their role mixture rates, detecting groups with rare mixture rate no long applies. Therefore, we think of the task as a change point detection problem and aim to detect the groups whose mixture rates change drastically from the previous time stamps. Compared with GLAD, we not only need to decide whether a group is anomaly or not, but also need to specify when the group appears anomalous. 
 
%Denoting role mixture rate for group $m$ as $\theta_m$, we define
%
%\begin{equation*} \label{eqn-score-static}
%\text{Anomaly Score}_{\text{GLAD}}  =   \| \theta_m - \mathbb{E}(\theta_m)\|
%\end{equation*}
%\begin{equation*} \label{eqn-score-dynamic}
%\text{Anomaly Score}_{\text{d-GLAD}}  =  \| \theta_m^{(t)} - \theta_m^{(t-1)}\|
%\end{equation*}

Even though we use slightly different definitions of group anomaly for the GLAD model and the d-GLAD model, the key ideas behind our definitions are the same. Both  definitions build upon the notion of role mixture rate, which essentially requires a precise inference of both the group membership and role identity for each individual in the group.

\section{GLAD$^0$}
\label{sec:glad0}
Suppose that we are given a social network with $N$ people. Each person $p$ has total of $A_p$ activities.  The point-wise activities data is $\mathbf{X}$ $= $ $\{ \mathbf{X}_{1}$,$ \mathbf{X}_{2}$,$\dots, \mathbf{X}_{N}\}$. The pair-wise communication data is  $\mathbf{Y}$  $=$  $\{ Y_{1,1}$,$ Y_{1,2}$, $\dots,Y_{N,N} \}$. $\mathbf{X}_{p}\in \mathbb{R}^{V\times A_p}$. For a particular activity $a$, $X_{pa}$ consists of $V$ entries, denoting a feature vector of $V$ dimensions. $Y_{p,q}\in \{0,1\}$ is a binary valued variable, indicating the pair-wise relationship of nodes. These two forms of data are our inputs. Our goal is to analyze these data jointly and declare the group that has irregular role mixture rate as anomaly. In the following sections, we first describe the motivation for our hierarchical Bayes model and provide its generative process and the plate notation. Then we derive the inference algorithm using the variational Bayesian approach.

\subsection{Model Specification}
\begin{figure}[htbp]
\begin{minipage}[b]{0.45\linewidth}
\centering
\includegraphics[scale = 0.4]{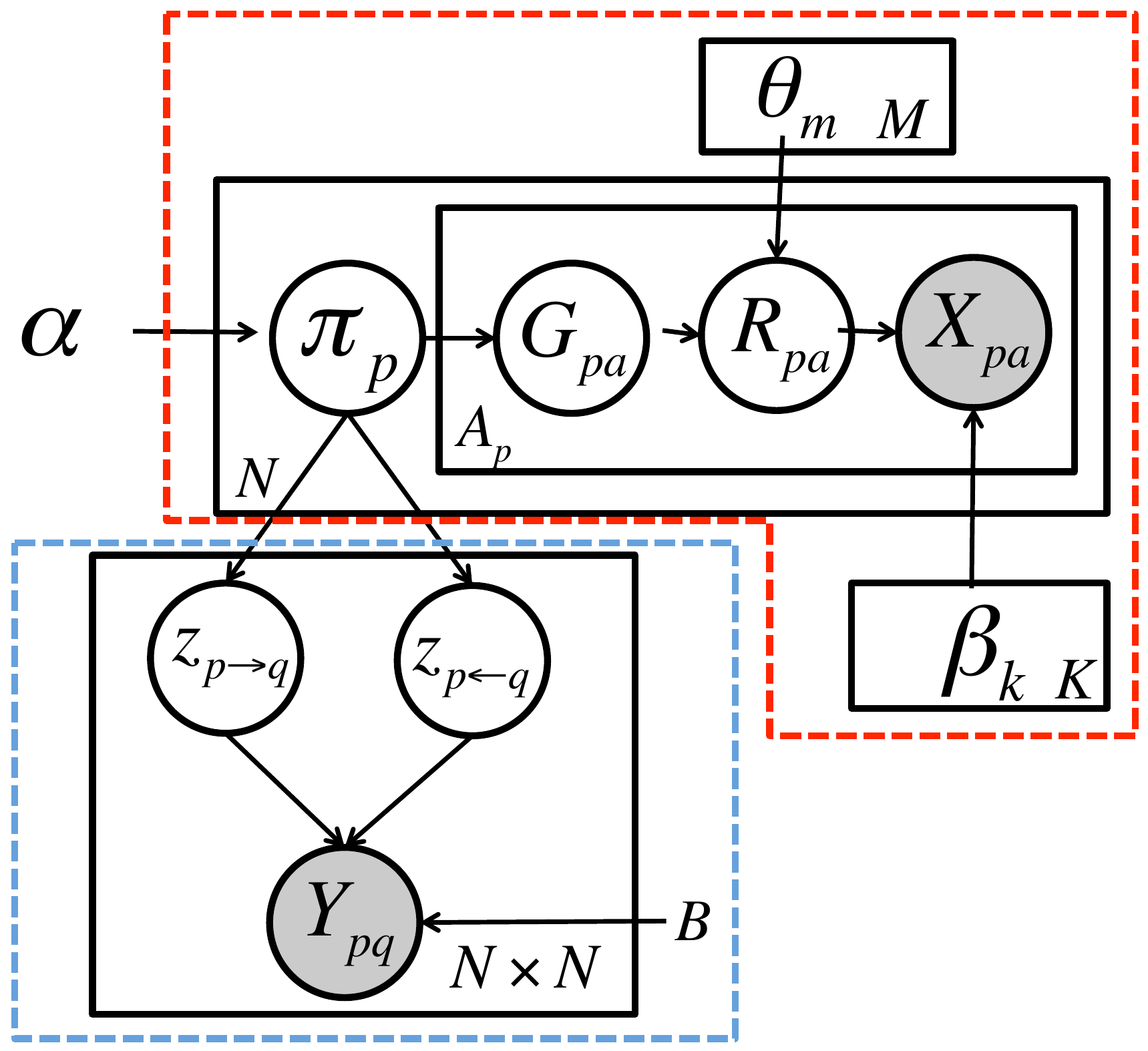}
\par\vspace{0pt}
\end{minipage}
\begin{minipage}[b]{0.47\linewidth}
\centering
\begin{tabular}{|c|c|} \hline
\texttt{Symbol} & \texttt{Description}\\ \hline
$\alpha$ & Dirichlet prior parameter \\ \hline
$\pi_p$ & group membership distribution of person $p$\\ \hline
$Y_{p q}$& pair-wise communication between $p$ and $q$ \\ \hline
$z_{p\rightarrow q}$& communication membership from $p$ to $q$\\ \hline
$z_{p \leftarrow q}$& communication membership  from $q$ to $p$\\ \hline
$B$&  global block probability among groups\\ \hline
$G_{pa}$& group identity of $p$ for activity $a$\\ \hline
$R_{pa}$& role identity of $p$  for activity $a$ \\ \hline
$\theta_{1:M}$& role mixture rate for M groups\\ \hline
$X_{pa}$ & activity $a$ of $p$ \\ \hline
$\beta_{1:K}$& activity distribution for K roles\\
\hline
\end{tabular}
\par\vspace{0pt}
\end{minipage}
\caption{Plate representation for the Group Latent Anomaly Detection (GLAD$^0$) model and the notation descriptions. Shaded circles are observations, blank circles are latent variables and the variables without a circle are model parameters. The blue rectangular resembles MMSB. The red polygon integrates the generating process of LDA. }
\label{fig:graphGLAD0}
\end{figure}

We model a social network with $N$ individuals.
From the point-wise data aspect, assume that each activity of the person $p$ is associated with a group identity  $G_{pa}$ and a role identity $R_{pa}$. Group identity finds the natural cluster of a person influenced by the pair-wise observations. Role identify captures the cluster of activities within the group.  The two identities assumption is motivated by the controversial viewpoints of what is the right metric for a community. In community detection literature \cite{ fortunato_community_2009}, some argue that a community is the one that has dense communications within clusters while others suggest that people in the same community should share common activity features. We get around the controversy by recognizing the arguments of both sides. Mathematically, since we model activities as a mixture model, ``role'' is the mixture component that categorizes the feature values of each activity. From the pair-wise data perspective, assume that each communication from person $p$ to $q$ has a group membership $z_{p\rightarrow q}$. The group membership of person $p$, $z_{p \rightarrow }$,  depends on the recipient
 of the communication while his group identity $G_p$ is undirected.  For simplification, we fix the number of groups as $M$ and the number of roles as $K$. 

For each person $p$, he joins a group according to the membership probability distribution $\pi_p$. We impose a Dirichlet prior on the membership distribution. It is well known that the Dirichlet distribution is conjugate to the multinomial distribution. As we will show later, when dealing with latent variables, the Dirichlet prior facilitates the learning and inference of the model. We assume the pair-wise link $Y_{p,q}$ between person $p$ and person $q$ depends on the group identities of both $p$ and $q$ with the parameter $B$. Furthermore, we model the dependency between the group and the role using a multinomial distribution parameterized by a set of role mixture rate $\{\theta_{1:M}\}$. The role mixture rate characterizes the constitution of the group: the proportion of the population that plays the same role in the group. Finally, we model the activity feature vector of the individual $X_{pa}$ as the dependent variable of his role with parameter set $\{\beta_{1:K}\}$.  Without loss of generality, we assume that the activity data has discrete value and follows the multinomial distribution of single trial,i.e, the categorical distribution. But we can easily adapt $X_{pa}$ to other form of activities.

Figure \ref{fig:graphGLAD0} shows the plate representation of the proposed model and summarizes the notations therein.  Our model unifies the ideas from both the Mixture Membership Stochastic Block (MMSB) model \cite{airoldi_mixed_2008} and the Latent Dirichlet Allocation (LDA) model \cite{blei_latent_2003}.  The blue dashed rectangular on the left side resembles MMSB which models the formation of groups using link information. The red dashed polygon integrates the generating process of LDA which is often used for topic extraction from documents.  We denote the current model design as GLAD$^0$ and specify the generative process of GLAD$^0$ in Algorithm \ref{alg:genGLAD}. Next, we describe the variational Bayes inference for the GLAD $0$ model.

\begin{algorithm}[h]
\caption{\textbf{Generative process of the GLAD$^{0}$ model}}\label{alg:genGLAD}
\begin{algorithmic}
\For{individual  $p = 1 \to N $}
\State{ Draw group membership distribution $\pi_{p}\sim\mbox{Dir}(\alpha)$}
\For{individual  $q = 1 \to N $}
\State{ Draw group membership  $z_{p\rightarrow q}\sim \text{Multinomial}(\pi_p)$}
\State{ Draw group membership  $z_{p\leftarrow q}\sim \text{Multinomial}(\pi_q)$}
\State{ Sample communication $Y_{p,q} \sim \text{Bernoulli  }(z_{p\rightarrow q}^TBz_{p\leftarrow q})$}
\EndFor

\For{activity $a =1 \to A_p $}
\State{ Draw group identity $G_{pa}\sim \text{Multinomial}(\pi_p)$}

\State{ Draw role identity $R_{pa}\sim \text{Multinomial}(R_{pa}|\theta_{1:M},G_{pa})$}
\State{ Sample activity $X_{pa}\sim \text{Multinomial}(X_{pa}|\beta_{1:K},R_{pa})$}
\EndFor

\EndFor
\end{algorithmic}
\end{algorithm}

\subsection{Model Inference}
\begin{algorithm}[h]
\caption{\textbf{Variational Inference for the alternative GLAD}}\label{alg:varGLAD}
\begin{algorithmic}
\State{randomly initialize $B$, $\theta$, $\beta$}
\State{normalize $\theta$, $\beta$}
\Repeat
\State initialize $\phi_{p\rightarrow q,g}:= 1/M$
\State initialize $\phi_{p\leftarrow q,h}:= 1/M$
\State initialize $\gamma_{p,g}:= 1/M$
\State initialize $\mu_{pa,r}:= 1/K$
\State initialize $\lambda_{pa,g}:= 1/M$
\Repeat
\For{ $p =1 \to N$}
\State{update $\gamma_{p,g} = \alpha_g + \sum_{q=1}^N \left[\phi_{p\rightarrow q,g}+\phi_{p\leftarrow q,g}\right] + \sum_{a=1}^{A_p}\lambda_{pa,g} $}
\For{$q = 1 \to N $, $g = 1 \to M$,  $h = 1 \to M$}
\State{update $\phi_{p\rightarrow q,g} \propto e^{\mathbb{E}_{q(\pi_p)}[\log\pi_{p,g}]}\cdot\prod_{h=1}^M\left[B_{gh}^{Y_{pq}}(1-B_{gh})^{1-Y_{pq}}\right]^{\phi_{p\leftarrow q,h}}$ }
\State{update $\phi_{p \leftarrow q,h}  \propto e^{\mathbb{E}_{q(\pi_p)}[\log\pi_{p,h}]}\cdot\prod_{g=1}^M\left[B_{gh}^{Y_{pq}}(1-B_{gh})^{1-Y_{pq}}\right]^{\phi_{p\rightarrow q,g}}$ }
\EndFor
\For{$a= 1\to A_p $,  $g = 1 \to M$, $r = 1 \to K$}
\State{update $\lambda_{pa,g} \propto e^{\psi(\gamma_{p,g})}\cdot\prod_{r=1}^K
\theta_{gr}^{\mu_{ga,r}}$}
\State{update $\mu_{pa,r}  \propto \prod_{g=1}^M\theta_{gr}^{\lambda_{pa,g}}\cdot\prod_{d=1}^D\beta_{rd}^{x_{pa,d}}
$}
\EndFor
\EndFor
\Until{convergence}
\State{ update $B_{gh} = \frac{\sum_{p,q}Y_{pq}\phi_{p\rightarrow q,g}\phi_{p\leftarrow q,h}}{(1-\rho)\cdot\sum_{p,q}\phi_{p\rightarrow q,g}\phi_{p\leftarrow q,h}}$ }
\State{ update $\beta_{rd}\propto \sum_p\sum_a x_{pa,d}\cdot\mu_{pa,r}$ }
\State{ update $\theta_{gr}\propto \sum_p\sum_a \lambda_{pa,g}\mu_{pa,r}$ }
\Until{convergence}

\end{algorithmic}
\end{algorithm}

We develop an approximate inference technique based on variational  Bayesian methods  \cite{jordan_introduction_1999} and an EM algorithm for model inference. Specially, we approximate analytically  to the posterior probability of the hiddent variables by minimizing the  Kullback - Leibler  divergence (KL-divergence) of the variational distribution and the actual posterior. Then we perform the EM procedure to learn the model parameters.

Denote the set of model parameters as  $\Theta=\{\alpha, B, \theta_{1:M},\beta_{1:K}\}$, the set of visible variables as $v= \{X_{1:N}, Y_{1:N,1:N}\}$, and the set of the hidden variables as $h$ $=$ $\{\pi_{1;N}$,$Z_{1:N,1:N}$,$ G_{1:N}$,$R_{1:N}\}$. Our aim is to estimate the posterior distribution $p(h,\Theta|v)$. We can first write out the complete joint likelihood of observed and latent variables as follows:
\begin{eqnarray*}\label{eqn:completeLikelihood}
p(v,h|\Theta) & = & \prod_{p,a}p(X_{pa}| R_{pa}, \beta) p (R_{pa}|G_{pa}, \theta) p(G_{pa}|\pi) \\
 & \times & \prod_{pq} p(Y_{pq}|z_{p\rightarrow q},z_{p\leftarrow q}  ) p(z_{p\rightarrow q}| \pi_p)p(z_{p\leftarrow q}| \pi_p)  \prod_p  p(\pi_p|\alpha).
\end{eqnarray*}

The marginal likelihood of the data $ p(v|\Theta) = \int_{h} p(v,h|\Theta)d h$ requires to integrate over all the latent variables in the equation above, which is intractable \cite{airoldi_mixed_2008}. Therefore, we choose a variational distribution $q(h)$ to approximate the actual posterior distribution, so that the Kullback-Leibler divergence (KL-divergence) between the actual posterior  $p(h|\Theta, v)$ and its approximation $q(h)$ is minimized.  Rewriting the marginal log likelihood and plugging in the variational distribution, we have
\begin{eqnarray*}
 \log p(v|\Theta) = D_{KL}(p||q) + E_q [\log p(v,h|\Theta)] -  E_q [\log q(h) ],
\end{eqnarray*}
where $E_q [f]$ represents the expectation of the function $f$ with respect to the distribution $q$. Since the marginal likelihood $\log p(v|\Theta)$ is invariant to the choice of $q$, minimizing the KL-divergence $D_{KL}(p||q)$ is equivalent to maximizing the last two terms $\langle\log p(v,h|\Theta)\rangle_q -  \langle\log q(h)\rangle_q$. In practice, we choose $q(h)$ to be factorized over the latent variables with free parameters $\Delta = \{\gamma_{1:N},\phi_{1:N,1:N}, \mu_{1:N},\lambda_{1:N}\} $ as follows:

\begin{displaymath}
\label{eqn:variationalLikelihood}
 q(h|\Delta)  =   \left[\prod_{p} q(\pi_p|\gamma_p)\right]\left[\prod_{p,q} q(z_{p\rightarrow q}|\phi_{p\rightarrow q})q(z_{p\leftarrow q}|\phi_{p\leftarrow q}) \right]\left[\prod_{p}\prod_{a} q(G_{pa}|\lambda_{pa})q(R_{pa}|\mu_{pa}) \right].
\end{displaymath}

Finding the optimal set of the variational parameters is equivalent to solving the following optimization problem subject to probability constraints:
\begin{eqnarray*}\label{eqn:objective}
\Delta^{\star} &=& \argmax\limits_{\Delta} \langle\log p(v,h|\Theta)\rangle_q -  \langle\log q(h|\Delta)\rangle_q  \nonumber\\
 & =&  \argmax\limits_{\Delta} L(v, h,\Theta, \Delta) .
\end{eqnarray*}
We follow an EM procedure to solve the problem above. We iteratively update the free parameters by taking the derivative of the Lagrange function of the objective $L$ over one parameter at a time given the value of others from the last iteration. Since $\{Y_{p,q}\}$ is symmetric, the objective function will result in a quadratic term with respect to $\lambda_p$. Taking the derivative over the variational parameter would not have a closed form solution. A simple workaround is by assuming constant probability for the generation of $\{Y_{p,p}\}$. We omit the tedious derivations and  only present the final update formulas of each of the free parameters, as shown in Algorithm \ref{alg:varGLAD}. For convenience, we denote  $f(Y_{p,q},B_{m,n})= Y_{p,q}\log B_{m,n}+(1-Y_{p,q}) \log (1-B_{m,n})$.

For the parameter estimation, we apply the empirical Bayes method on the variational likelihood.  We maximize the Lagrange function of $L(v, h,\Theta, \Delta)$ over model parameters $\Theta=\{\alpha, B, \theta_{1:M},\beta_{1:K}\}$. Due to the fact that the derivative of the objective function with respect to $\alpha $ depends on $\alpha$, there is no closed form solution for the maximizer w.r.t $\alpha$. We apply the Newton-Raphson method to reach a numerical solution. Similar to the GLAD model, we score the group anomalousness using $-\sum_{p \in G}E_p[\log p(R_p|\Theta)]$. The most anomalous group will have the highest anomaly score. We approximate the true log likelihood with the variational log likelihood to get $-\sum_{p \in G} E_q [ \log p(R_p|\Theta)]$.

GLAD$^0$ jointly models the point-wise and pair-wise data. It allows mixture of groups and roles by associating each activity with a group identity and a role identity, which implies that each person can have multiple roles and can belong to multiple groups. The GLAD$^0$ model loosely connects the two components of MMSB and LDA via a shared group distribution $\pi_p$. It distinguishes between the communication group membership $z$ and the activity membership $G$.  However, the number of latent variables in GLAD$^0$ scales linearly with number of activities for each person, thus GLAD$^0$ suffers from high computational cost. The complexity  of the model and the difficulty of inference increase significantly when  we further consider generalizing to the dynamic setting. Additionally, the loose connection with the shared group membership $\pi_p$ may be restrictive in capturing the inter-dependencies of point-wise and pair-wise data. Therefore, we consider a more computationally efficient model design that addresses the above issues.

\section{GLAD}
\label{sec:glad}
\begin{figure}[t]
\begin{minipage}[b]{0.45\linewidth}
\centering
\includegraphics[scale = 0.8 ]{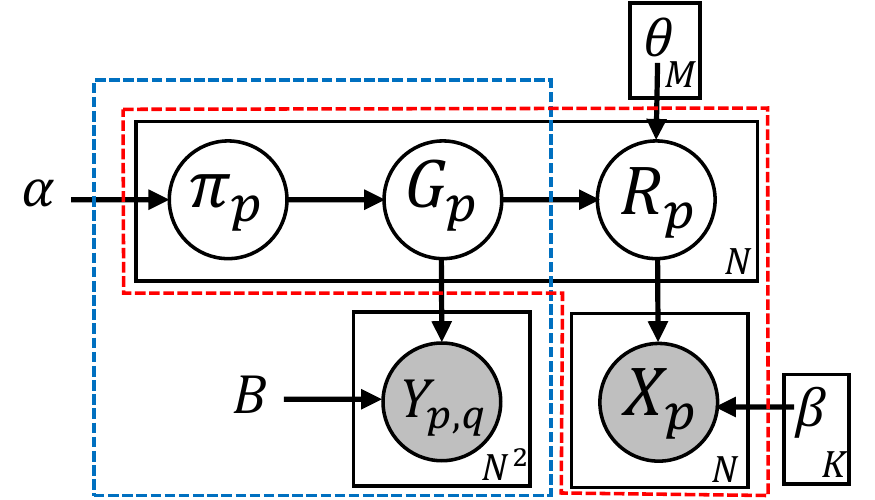}
\end{minipage}
\begin{minipage}[b]{0.47\linewidth}
\centering
\begin{algorithmic}
\For{individual  $p = 1 \to N $}
\State{ Draw membership distribution $\pi_{p}\sim\mbox{Dir}(\alpha)$}
\State{ Draw $G_{p}\sim \text{Multinomial}(\pi_p)$}
\For{individual  $q = 1 \to N $}
\State{Sample $Y_{p,q} \sim \text{Bernoulli  }(G_{p}^TBG_{q})$}
\EndFor
\State{ Draw $R_{p}\sim \text{Multinomial}(R_p|\theta_{1:M},G_p)$}
\State{ Draw $X_{p}\sim \text{Multinomial}(X_p|\beta_{1:K},R_p)$}
\EndFor
\end{algorithmic}
\end{minipage}
\caption{Plate notation for the GLAD model and the corresponding generative process. Shaded circles are observations, blank circles are latent variables and the variables without a circle are model parameters.}
\label{fig:graphGLAD}
\end{figure}

%%%%%%%%%%%%%%%%%%%%%%%%%%%%%%%%%%%%%%%%%%%%%%%%%%%%%%%%%
GLAD models a social network of activities $\mathbf{X}$ $= $ $\{ X_{1}$,$ X_{2}$,$\dots, X_{N}\}$ and communications $\mathbf{Y}$  $=$  $\{ Y_{1,1}$,$ Y_{1,2}$, $\dots,Y_{N,N} \}$, where $X_{p}$ is the aggregation of the activities for each person. $X_{p}\in \mathbb{R}^V$ consists of $V$ entries, denoting a feature vector of $V$ dimensions.  Each person $p$ joins a group according to the membership probability distribution $\pi_p$. He is associated with a group identity $G_p$ and a role identity $R_p$.  We draw the pair-wise observations of person $p$ $\{Y_{p,:}\}$ directly from the group identity $G_p$ as Bernoulli random variables. And we further assume that the activities $X_{p}$ follows a multinomial distribution with $A_p$ trials. GLAD incorporates MMSB and LDA in a more compact way. It not only allows the shared group membership distribution  between the two components, but also the group membership identity to emphasize the inter-dependencies between point-wise and pair-wise data. Figure \ref{fig:graphGLAD} depicts the plate representation of the GLAD model and its corresponding generative process. 

\subsection{Inference \lowercase{and} Learning}
Inference requires us to compute the posterior distributions of the latent variables given the data. The normalizing term of the posterior distribution involves the calculation of the marginal likelihood of the data for which we resort to variational EM algorithms \cite{jordan_introduction_1999}.

Denote the set of model parameters as  $\Theta=\{\alpha, B, \theta_{1:M},\beta_{1:K}\}$, the set of observed variables as $v= \{X_{1:N}, Y_{1:N}\}$, and the set of the hidden variables as $h$ $=$ $\{\pi_{1;N}$,$ G_{1:N}$,$R_{1:N}\}$. Our aim is to estimate the posterior distribution $p(h,\Theta|v)$. We can first write out the complete joint likelihood of observed and latent variables as follows:
\begin{eqnarray*}\label{eqn:completeLikelihood}
p(v,h|\Theta) & = & \prod_{p}p(\pi_p|\alpha) \times
\prod_{p,q}p( Y_{p,q}|G_{p}, G_{q},B) \\
& \times &  \prod_{p}p(X_p|R_p, \beta_{1:K})p(R_p|G_p,\theta_{1:M})p(G_p|\pi_p).
\end{eqnarray*}

Computing the maximizer for the marginal likelihood of the data $ p(v|\Theta) = \int_{h} p(v,h|\Theta)d h$ requires the integration over all the latent variables in the equation above, which is intractable \cite{airoldi_mixed_2008}. Therefore, we apply the variational Bayesian approach \cite{jordan_introduction_1999} to perform the inference approximately. The essence of the variational Bayesian approach is to choose a variational distribution $q(h)$ to approximate the actual posterior distribution, so that the Kullback-Leibler divergence (KL-divergence) between $p(h,\Theta|v)$ and its approximation $q(h)$ is minimized.

Rewriting the marginal log likelihood and plugging in the variational distribution, we have
\begin{eqnarray*}
 \log p(v|\Theta) = D_{KL}(p||q) + \langle\log p(v,h|\Theta)\rangle_q -  \langle \log q(h)\rangle_q ,
\end{eqnarray*}
where we use $\langle f \rangle_p$ to represent the expectation of the function $f$ with respect to the distribution $p$. Since the marginal likelihood $\log p(v|\Theta)$ is invariant to the choice of $q$, minimizing the KL-divergence $D_{KL}(p||q)$ is equivalent to maximizing the last two terms $\langle\log p(v,h|\Theta)\rangle_q -  \langle\log q(h)\rangle_q$. In practice, we choose $q(h)$ to be factorized over the latent variables with free parameters $\Delta = \{\gamma_{1:N},\mu_{1:N},\lambda_{1:N}\} $ as follows:

\begin{displaymath}
\label{eqn:variationalLikelihood}
 q(h|\Delta)  =  \prod_p q (\pi_p|\gamma_p) q(R_p|\mu_p) q(G_p|\lambda_p) .
\end{displaymath}

Our goal is to find the optimal set of free parameters that provides a variational distribution closest to the actual posterior. Then our problem is to maximize the objective function formulated as follows subject to probability constraints:

\begin{eqnarray*}\label{eqn:objective}
\Delta^{\star} &=& \argmax\limits_{\Delta} \langle\log p(v,h|\Theta)\rangle_q -  \langle\log q(h|\Delta)\rangle_q  \nonumber\\
 & =&  \argmax\limits_{\Delta} L(v, h,\Theta, \Delta) .
\end{eqnarray*}

The objective function $L$, by plugging in the joint likelihood and the variational distribution and taking expectations, is given by
\small
\begin{eqnarray*}\label{eqn:logLikelihood}
L(v, h,\Theta, \Delta)  & = &  \sum_{p}\langle\log p(X_p|R_p, \beta_{1:K})\rangle_q \nonumber  +  \sum_{p} \langle\log p(R_p|G_p,\theta_{1:M})\rangle_q  + \sum_{p} \langle\log p(G_p|\pi_p)\rangle_q   \\
 &+ & \sum_{p,q} \langle\log p( Y_{p,q}|G_p, G_q,B)\rangle_q  + \sum_{p} \langle\log p(\pi_p|\alpha)\rangle_q  \\
& - & \sum_p \langle\log q (\pi_p|\gamma_p)\rangle_q   
 -  \sum_p \langle\log q(R_p|\mu_p)\rangle_q  - \sum_p \langle\log q(G_p|\lambda_p)\rangle_q .
\end{eqnarray*} 
\normalsize

We follow a variational EM procedure in order to maximize  $L(v, h,\Theta, \Delta)$ over $\Delta$. Basically we iteratively update the free parameters by taking the derivative of the Lagrange function of the objective $L$ over one parameter at a time given the value of others from the last iteration. Since $\{Y_{p,q}\}$ is symmetric, the objective function will result in a quadratic term with respect to $\lambda_p$. Taking the derivative over the variational parameter would not have a closed form solution. A simple workaround is by assuming constant probability for the generation of $\{Y_{p,p}\}$. We omit the tedious derivations and  only present the final update formulas of each of the free parameters, as shown in Algorithm \ref{alg:varGLAD}. For convenience, we denote  $f(Y_{p,q},B_{m,n})= Y_{p,q}\log B_{m,n}+(1-Y_{p,q}) \log (1-B_{m,n})$.

\begin{algorithm}[h]
\caption{\textbf{Variational Inference for GLAD}}\label{alg:varGLAD}
\begin{algorithmic}
\State initialize $\gamma_{p,m}:= 1/M$
\State initialize $\mu_{p,k}:= 1/K$
\State initialize $\lambda_{p,m}:= 1/M$
\Repeat
\For{ $p =1 \to N$ , $m = 1 \to M$  $k = 1 \to K$}
\State{$\gamma_{p,m} = \alpha_m +\lambda_{p,m}$}
\State{$\lambda_{p,m}=\exp \{\sum_k \log\theta_{m,k}\mu_{p,k}+\psi(\gamma_{p,m})-\psi(\sum_n\gamma_{p,n})
+ \sum_{q\neq p}\sum_n \lambda_{q,n} \cdot f(Y_{p,q},B_{m,n})\}$}
\State{$\mu_{p,k}=\exp \{ \sum_v \log\beta_{v,k} X_{p,v}+\sum_m \log\theta_{m,k}\lambda_{p,m} \}$}
\EndFor
\Until{convergence}

\end{algorithmic}
\end{algorithm}

For the parameter estimation, we apply the empirical Bayes method on the variational likelihood.  We maximize the Lagrange function of $L(v, h,\Theta, \Delta)$ over model parameters $\Theta=\{\alpha, B, \theta_{1:M},\beta_{1:K}\}$. We apply the Newton-Raphson method to reach a numerical solution for the maximizer w.r.t $\alpha$. The resulting parameter updating functions for $\alpha$ and $B$ are the same as those of MMSB \cite{airoldi_mixed_2008} and the parameters $\beta$ and $\theta$ can be estimated as follows:

\begin{equation*}
\beta_{v,k} = \frac{\sum_p X_{p,v}\mu_{p,k}}{\sum_{v,p} X_{p,v}\mu_{p,k}}
\quad
\theta_{m,k}= \frac{\sum_p \mu_{p,k}\lambda_{p,m}}{\sum_{k,p}\mu_{p,k}\lambda_{p,m}}.
\end{equation*}

\begin{figure*}[t]
\centering
\begin{minipage}[c]{80mm}
\includegraphics[width = 85mm]{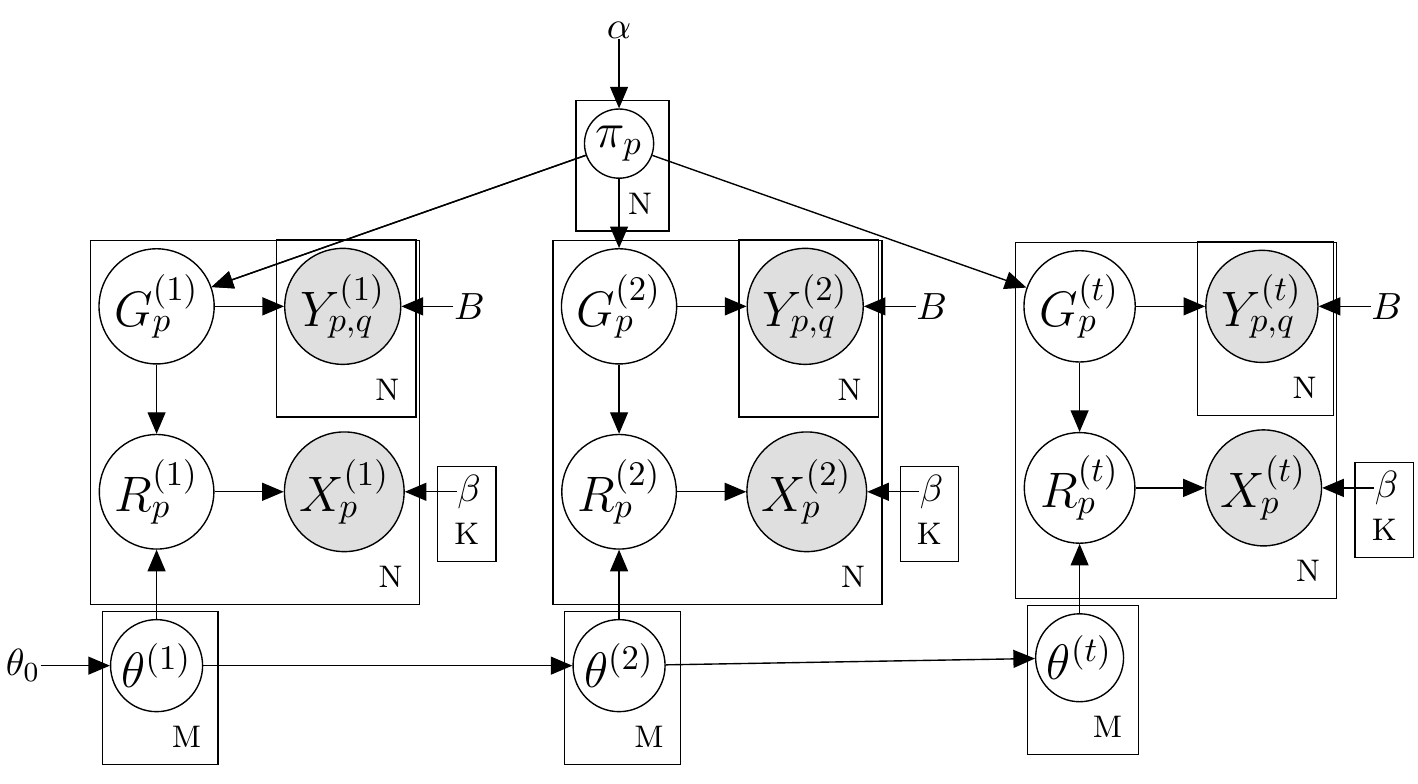}
\end{minipage}
\qquad
\footnotesize
\begin{tabular}{|c|c|} \hline
\texttt{Notation} & \texttt{Description}\\ \hline
$\alpha$ & Dirichlet parameter\\ \hline
$\pi_p$ & membership distribution \\ \hline
$G_{p}^{(t)}$& group of $p$ at time $t$\\ \hline
$Y_{p,q}^{(t)}$& pair-wise communication at time $t$  \\ \hline
$B$& global block probability \\ \hline
$R_{p}^{(t)}$& role of $p$ at time $t$ \\ \hline
$\beta_{1:K}$& activity mixture rate\\ \hline
$X_{p}^{(t)}$& activity of $p$ at time $t$ \\ \hline
$\theta_0$ & initial Gaussian mean  \\ \hline
$\theta_{1:M}^{(t)}$& role mixture rates at time $t$ \\ \hline
\end{tabular}
\caption{Plate notation for the d-GLAD model and the meaning of notations. The subscript $p$ denotes each person in the social network. The superscript $t$ denotes the network snapshot at time stamp $t$. }
\label{fig:graphDGLAD}
\end{figure*}

We score the group anomalousness using $-\sum_{p \in G}\langle\log p(R_p|\Theta)\rangle_p$ according to our definition of group anomaly in section \ref{sec:definition}. The most anomalous group will have the highest anomaly score. We approximate the true log likelihood with the variational log likelihood to get $-\sum_{p \in G} \langle\log p(R_p|\Theta)\rangle_q$.

A limitation of GLAD is that it only models the static network. This might be restrictive if we want to further consider dynamic networks. Besides the anomaly group whose mixture rate deviates significantly from other groups, we are also interested to study how the mixture rate evolves over time. Fortunately, GLAD can be easily extended to account for this dynamics. This leads to the dynamic extension of the GLAD model, which will be discussed in the next section.

\section{dynamic GLAD}
\label{sec:dglad}
%dglad
We now generalize the GLAD model to take into account the dynamics in the social media. We refer the dynamic extension of GLAD as the d-GLAD model. To be consistent with our description for GLAD in section \ref{sec:glad}, we start with the model specification and then provide the model inference algorithm using both the variational Bayesian method and the Monte Carlo sampling technique.

\subsection {Model Specification}
Generalization of GLAD to d-GLAD stems from the template models \cite{koller2009probabilistic}, which use the model for a particular time stamp as a template, duplicate it over time and connect temporal components sequentially. Similarly, we can adapt GLAD to the dynamic setting by making a copy of GLAD for each time point. To simplify the model, we assume that the latent factors including role $R_p$, group $G_p$ and mixture rate $\{\theta_{1:M}\}$ change over time but the membership distribution $\{\pi_p\}$ and model parameters are fixed. 

We model the temporal evolution of the role mixture rate for each group with a series of multivariate Gaussian distributions. At a particular time point, the Gaussian has its mean as the value of the mixture rate. And the mixture rate of the next time point is a normalized sample from this Gaussian distribution. Since we require the mixture rate to be the parameters of a multivariate distribution over features, we apply a soft-max function to normalize the sample drawn from the multivariate Gaussian. The soft-max function is defined as $S(\theta_m) = \frac{\exp \theta_m}{\sum\limits_m \exp \theta_m}$. When the total time length $T$ equals one, d-GLAD reduces to the GLAD model. Figure \ref{fig:graphDGLAD} depicts the probabilistic graphical model of d-GLAD and the meanings of notations used. We summarize the generative process of d-GLAD in Algorithm \ref{alg:genDGLAD}.

In d-GLAD model, since the mixture rate of next time stamp is drawn from a multivariate Gaussian centering around the mixture rate of its previous time stamp, it imposes smoothness on the mixture rates across time, preventing the mixture rate from having drastic changes. The soft-max function maps the samples from the multivariate Gaussian to the parameters for the multinomial distribution. Similar idea can be seen from the generalization of LDA to the dynamic topic model \cite{blei2006dynamic}. While it is true that d-GLAD model shares the constraints of GLAD on fixed group/role number and constant self-loop, it has certain intriguing advantages over static models. (\romannumeral 1)
d-GLAD captures the dynamics of the latent variables $G_p$ and $R_p$, thus allows an individual to switch groups and roles over time.(\romannumeral 2) The smoothness of the mixture rate over time models the behavior of normal groups, so detecting groups whose mixture rates $\theta_m^t$ undergo substantial change becomes easier.

\begin{algorithm}[h]
\caption{\textbf{Generative Process of DGLAD}}
\begin{algorithmic}
\For { $t = 1 \to T$}
\For { $m = 1 \to M$}
\State{ Draw $\theta_{m}^{(t)} \sim \text{Gaussian}(\theta_{m}^{(t-1)},\sigma)$}
\EndFor

\For {individual $p = 1 \to N$}
\State{ Draw membership distribution $\pi_{p}^{(t)}\sim\mbox{Dir}(\alpha)$}

\State{ Draw $G_{p}^{(t)}\sim \text{Multinomial}(\pi_p)^{(t)}$}
\For{individual  $q = 1 \to p-1 $ and $ q = p+1 \to N$  }
\State{Sample $Y_{(p,q)}^{(t)} \sim \text{Binomial }((G_{p}^{(t)})^TBG_{q}^{(t)})$}
\EndFor

\State{ Draw $R_{p}^{(t)}\sim \text{Multinomial}(R_p^{(t)}|\mathcal{S}(\theta_{G_p^{(t)}}^{(t)}))$}
\State{ Draw $X_{p}^{(t)}\sim \text{Multinomial}(X_p^{(t)}|\beta_{R_p^{(t)}})$}
\EndFor
\EndFor
\end{algorithmic}
\label{alg:genDGLAD}
\end{algorithm}

\subsection{Inference \lowercase{and} Learning}
The variational inference of d-GLAD is similar to the GLAD model except for the longitudinal factor $\theta_{1:M}^{(1:T)}$. We add a variational distribution $p(\theta_{m}^{1:T}|\hat{\theta}^{1:T})$ to approximate the original posterior where $\{\hat{\theta}^{1:T}\}$ are variational parameters. Then we apply the variational Kalman Filter technique \cite{blei2006dynamic}
to infer the sequential latent variables and learn the model parameters.
The transition for the  mixture rate of each group is Gaussian distributed:
\begin{equation*}
\theta^{(t)}|\theta^{(t-1)} \sim \mathcal{N}(\theta^{(t-1)}, \sigma^2 I).
\end{equation*}

We can write the variational distribution for the transition as follows:
\begin{equation*}
\hat{\theta}^{(t)}|\theta^{(t-1)} \sim \mathcal{N}(\theta^{(t-1)}, \hat{v}^2 I).
\end{equation*}

Then we can apply similar variational EM procedure incorporating the transitions to infer the variational parameters.
Due to the numerical difficulty of variational Kalman filter method, we also implement a version of the Monte Carlo sampling for d-GLAD model, which is used in our empirical evaluations. The algorithm is elaborated in Algorithm \ref{alg:sampleDGLAD}. The inference of the transitional part $\{\theta^{1:T}\}$ is based on the Particle Filtering method \cite{doucet2009tutorial}. The anomaly score of the d-GLAD model is measured by $\| \theta_m^{(t)} - \theta_m^{(t-1)}\|$.
\begin{algorithm}[h]
\caption{\textbf{Monte Carlo Sampling of DGLAD}}
\begin{algorithmic}
\State{ Initialize $\alpha , \theta_0,\beta_{1:K}, B $}
\State { $R_{1:N}^{1:T} = 1/K$, $G_{1:N}^{1:T}= 1/M$ , $\pi_{1:N} \sim \text{Dir} (\alpha)$ }
\Repeat{}
\For { $p = 1 \to N$}
\For { $t = 1 \to T $}
\State{ Update $R_{p}^{(t)}\sim \text{Mul}(\mathcal{S}(\theta_{G_p^{(t-1)}}^{(t-1)})) \text{Mul}(X_p^{(t)})$}
\State{ Update $G_{p}^{(t)}\sim \text{Mul}(\pi_p^{(t-1)}) \text{Mul}(\mathcal{S}(\theta_{G_p^{(t-1)}}^{(t-1)})) $}
\EndFor
\State{ Update $\pi_{p} \sim \text{Dir} (\alpha) $ }
\EndFor
\For { $t = 1 \to T $}
\State{ Update $\theta^{(t)} $ using Particle Filtering}
\EndFor
\Until {Convergence}
\end{algorithmic}
\label{alg:sampleDGLAD}
\end{algorithm}

\section{Experiments}
\label{sec:experiment}
% updated experiment session
To evaluate the effectiveness of our model, we conduct thorough experiments on synthetic datasets and real world datasets. We study the applications of our approach by analyzing scientific publications and senator voting records.

\subsection{Baselines}
To our knowledge, all existing algorithms are two-stages approaches: (\romannumeral 1) identify groups, (\romannumeral 2) detect group anomalies. We summarize these algorithms in Table \ref{two_step_summary}. We use following approaches as baseline methods in comparison to GLAD and d-GLAD:

\begin{enumerate}
\item\textbf{MMSB-LDA}:First use the MMSB model to learn a group membership distribution for each individual node, then assign the node to the group with the highest probability. Finally, for each group, train an LDA model and infer the role identity.
\item \textbf{MMSB-MGM}: Group is learned using the same method as MMSB-LDA. For the role inference,  train an multi-modal MGM instead of LDA.
\item \textbf{Graph-LDA}: Run an off-the-shelf graph clustering algorithm Min-Cut to get group membership and then train a LDA model for each group.
\item \textbf{Graph-MGM}: Get group membership with the graph clustering algorithm Min-Cut and then train a MGM model for each group.
\end{enumerate}

\begin{table}[h]
\centering
\caption{Two stage models in existing work}
\begin{tabular}{|c|c|c| } \hline
\texttt{Algorithm} & \texttt{Stage-1} & \texttt{Stage-2}\\ \hline
Heard 2010 \cite{heard_bayesian_2010}& spectrum  & Poisson process \\ \hline
Xiong 2011-a \cite{xiong_hierarchical_2011} & clustering & Mixture Genre Model \\ \hline
Xiong 2011-b \cite{xiong_group_2011}& clustering & Flexible Genre Model  \\ \hline
Muandet 2013  \cite{muandet2013one} & simulator & One class SMM \\
\hline\end{tabular}
\label{two_step_summary}
\end{table}

\subsection{Synthetic Dataset}
We experiment on two type of synthetic datasets. One is a synthetic dataset with injected group anomalies. The other is a benchmark dataset generated by a simulator with individual anomaly labels.

\subsubsection{Synthetic Data with Anomaly Injection}
We generate a network with 500 nodes using GLAD in Algorithm \ref{alg:genGLAD}. To evaluate the anomaly detection performance, we set the mixture rates of anomalous groups as $[0.9, 0.1]$ and normal groups as $[0.1,0.9]$. We vary the number of groups from 5 to 50 and inject $20\%$ anomalous groups. The rest $80\%$ groups are normal. Since we know the normal and anomalous mixture rates, we calculate the anomaly score of each group by directly computing the differences between the inferred mixture rate and the ground truth normal mixture rate. During the testing procedure, we rank the groups with respect to their anomaly score and retrieve top $20\%$ groups.  For all methods, we set the number of groups and number of roles the same as the ground truth.
\begin{figure}[t]
\centering
\subfigure[Original]{
\includegraphics[scale = 0.17]{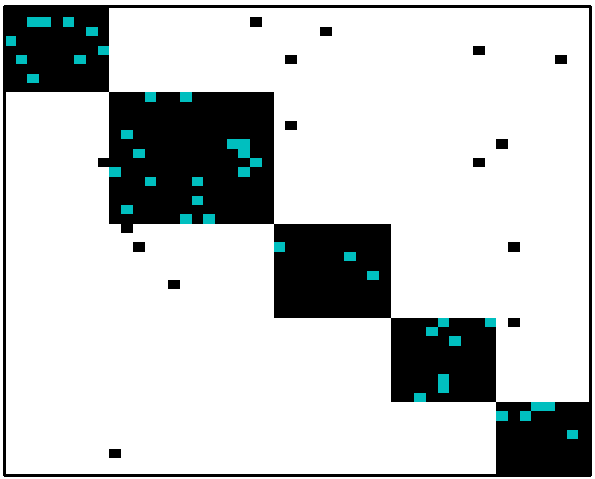}\label{fig:blockTruth}
}
\subfigure[GLAD]{
\includegraphics[scale = 0.17]{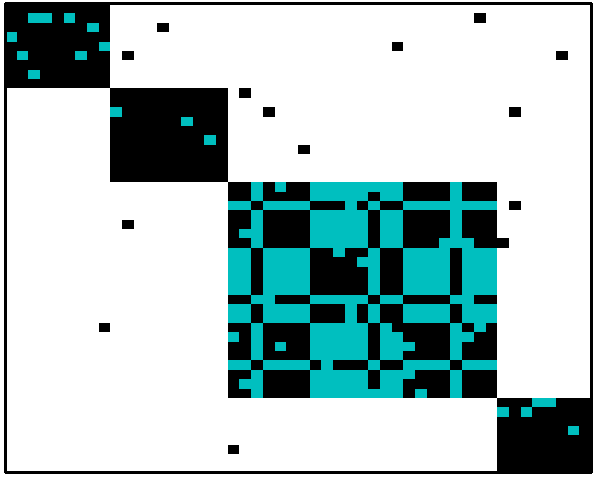}\label{fig:blockglad}
}
\subfigure[MMSB]{
\includegraphics[scale = 0.17]{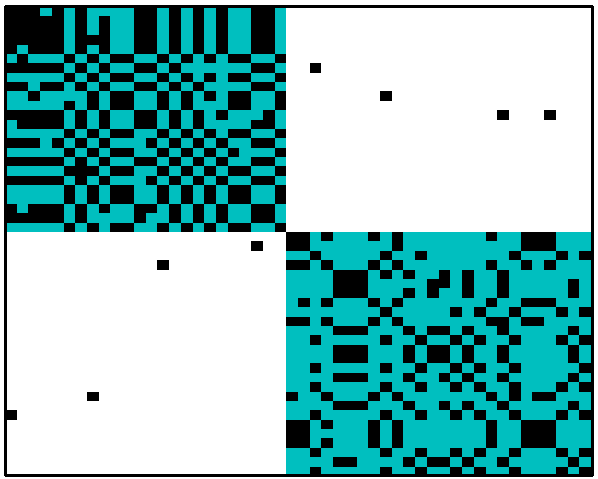}\label{fig:blockTruth}
}
\subfigure[Graph]{
\includegraphics[scale = 0.17]{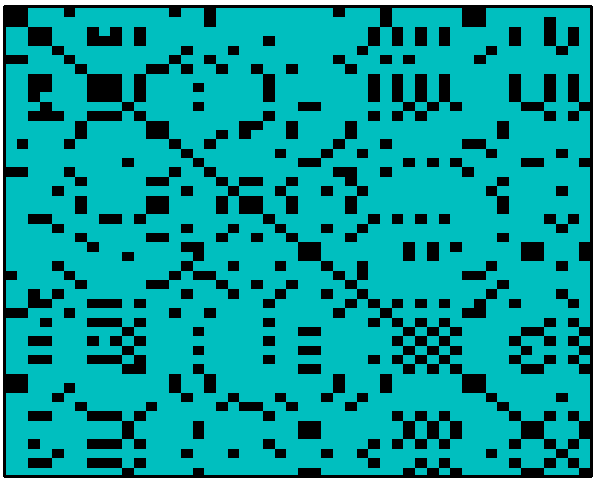}\label{fig:blockGraph}
}
\caption{The $50 \times 50$ adjacent matrix re-arranged by the group membership discovered  by three grouping approaches on a subset of synthetic data of 5 groups. Dark pixels denote links and white pixels denote no links. Blue block highlights the learned group membership. }
\label{fig:syn_block}
\end{figure}

\begin{figure*}[t]
\centering
\subfigure[Robustness]{
\includegraphics[scale= 0.23 ]{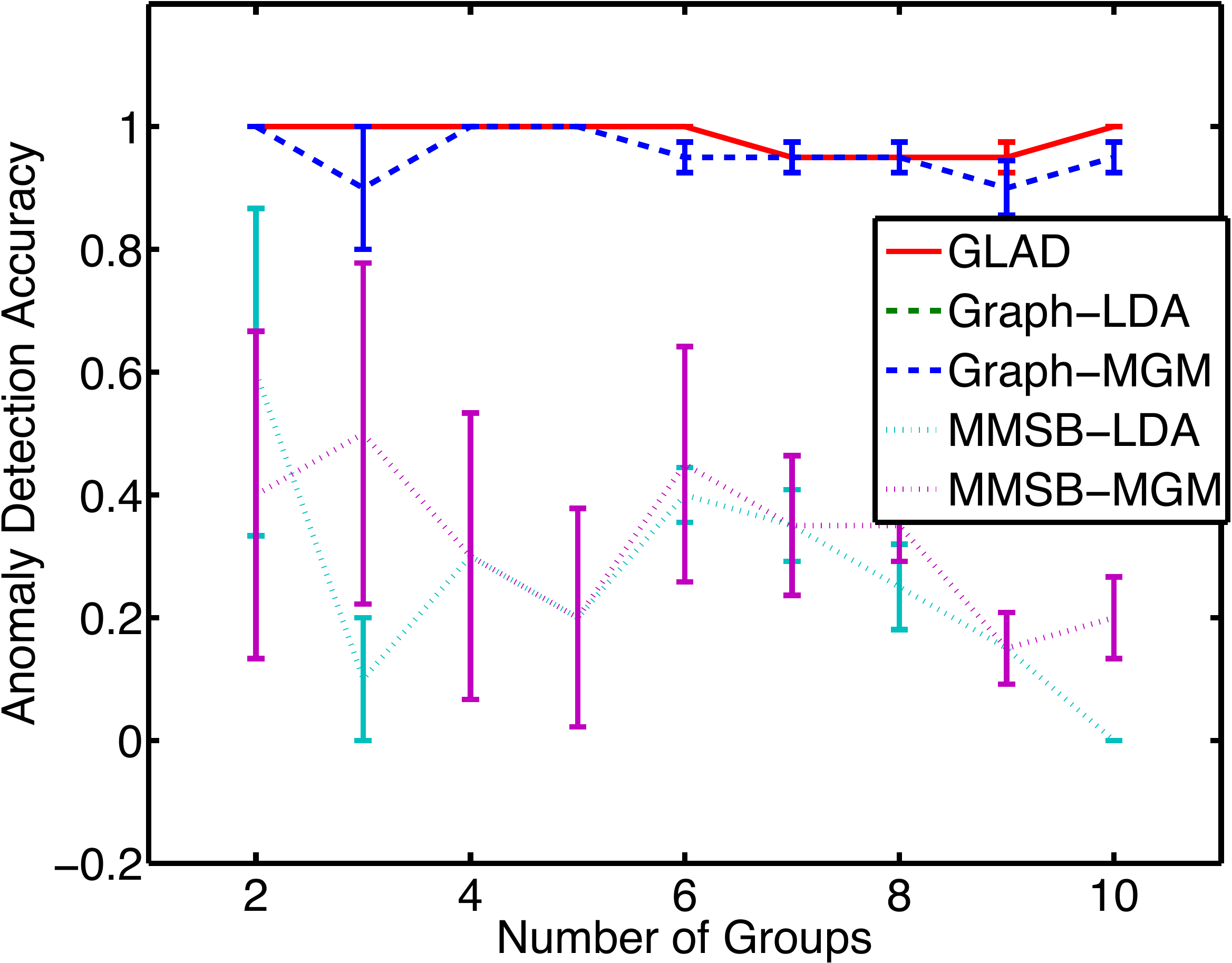} \label{fig:glad_robustness}
}
\subfigure[Accuracy]{
\includegraphics[scale= 0.23 ]{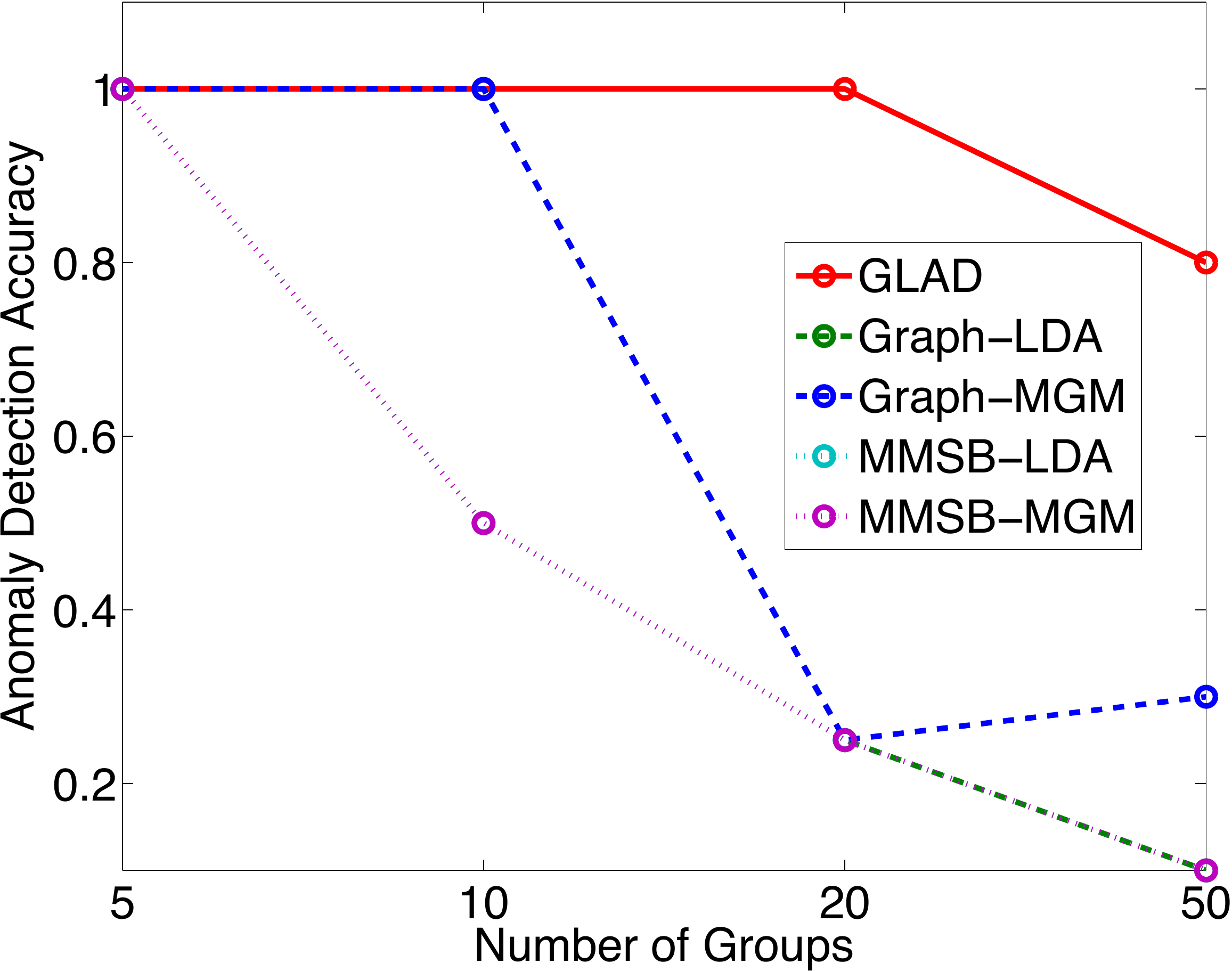} \label{fig:glad_accuracy}
}
\subfigure[GLAD$^0$]{
\includegraphics[scale= 0.23 ]{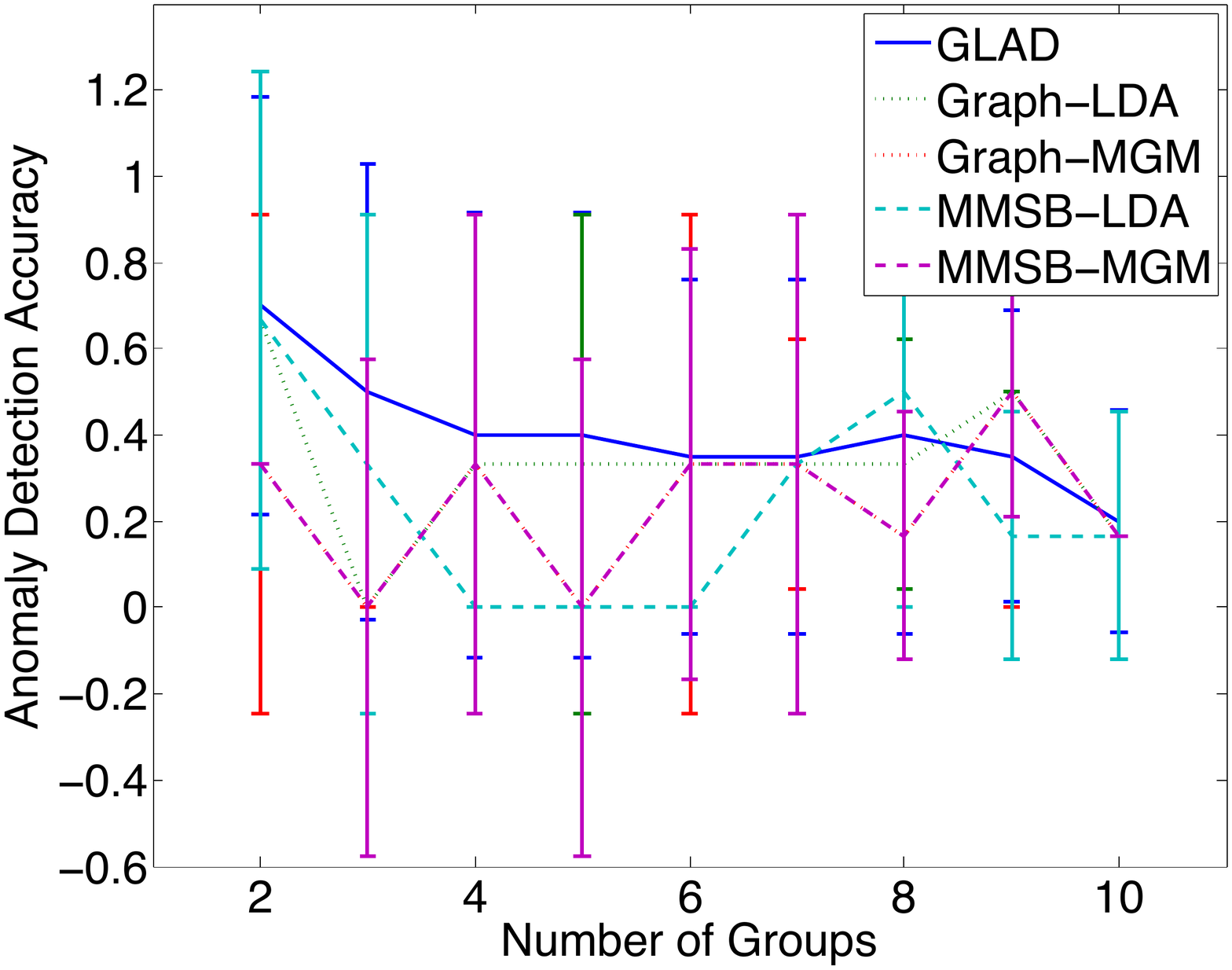} \label{fig:glad0_robustness}
}
\subfigure[d-GLAD]{
\includegraphics[scale = 0.23]{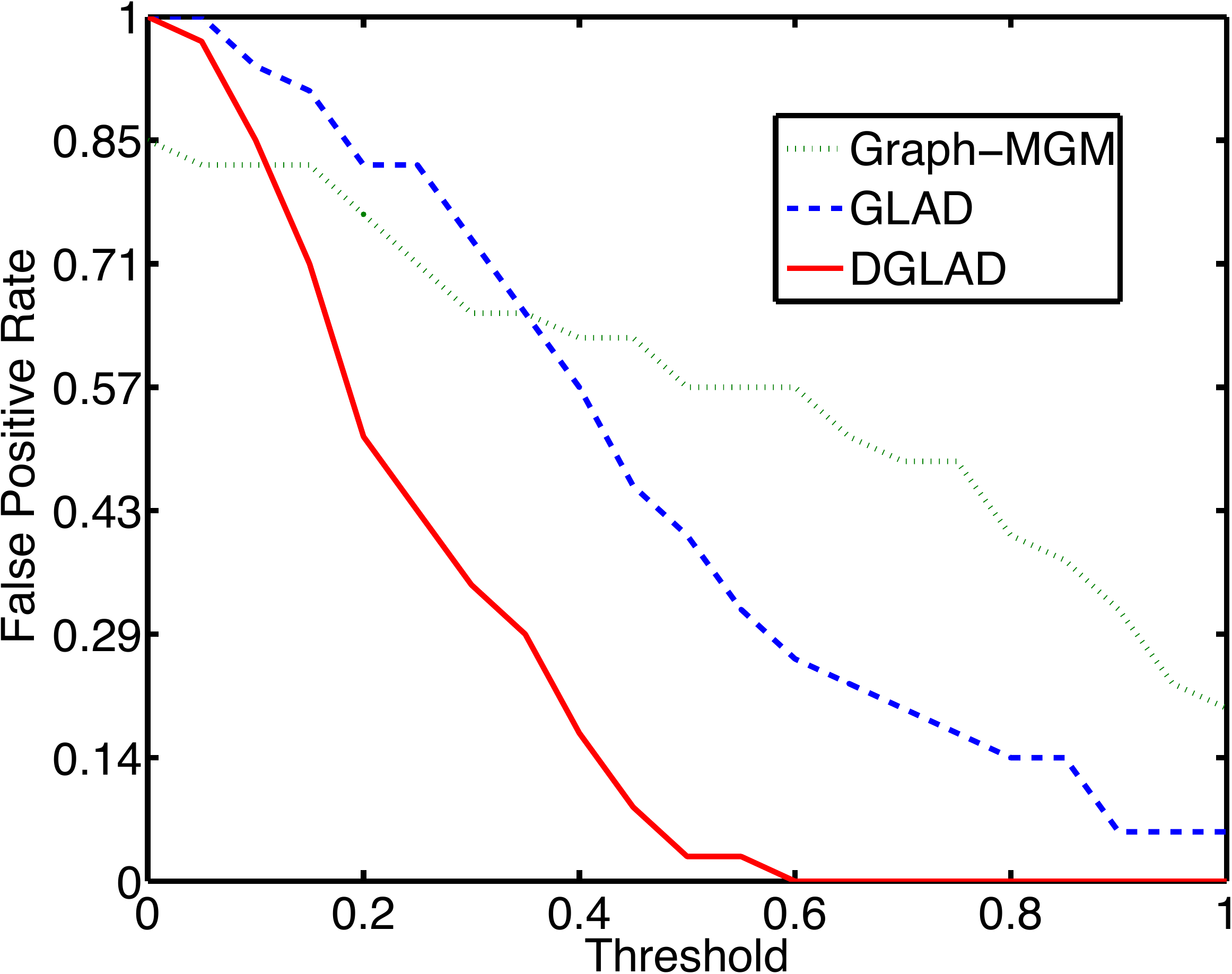}\label{fig:dglad_fp}
}
\caption{Anomaly detection performance of GLAD and baseline methods on synthetic dataset of 500 samples, with $20\%$ anomalous groups. \subref{fig:glad_robustness}: Detection accuracy  error bar plot of GLAD and four baselines for detection accuracy with group number from 2 to 10.  \subref{fig:glad_accuracy}: Mean detection accuracy for different number of groups up to 50, averaged over 10 random runs. 
\subref{fig:glad0_robustness}: Detection accuracy  error bar plot of GLAD$^0$ and four baselines for detection accuracy with group number from 2 to 10. \subref{fig:dglad_fp}:False positive rate over different thresholds for d-GLAD, MMSB-MGM and GLAD for synthetic data. $10\%$ group anomalies are injected. }
\label{fig:syn_anomaly}
\end{figure*}

We compare the learned groups of three grouping approaches with the ground truth: GLAD, MMSB and Graph, for the case of 5 groups. The inferred group memberships are shown as adjacent matrices in Figure \ref{fig:syn_block}. For better visualization, we intentionally put the nodes that belong to the same group together. Ideally, we should observe dense links within groups and sparse links between groups. Therefore, the dark pixels in the plot would aggregate along the principal diagonal of the matrix. We use blue color to highlight the groups learned. The group discovery result of GLAD is the closest to the ground truth. The high connectivity in the graph and the lack of point-wise information could be the reasons for the poor performance of Graph and MMSB.

Figure \ref{fig:glad_robustness} and  Figure  \ref{fig:glad_accuracy} shows the anomaly detection performance with different number of groups for GLAD and four other baselines. GLAD achieves the highest detection accuracy. It is also more robust over 10 random runs. Note that the differences for the first stage of baselines are more obvious than the second stage. This is because the Bernoulli distribution limits the number of samples in the pair-wise data, making the first stage more difficult to learn.

We justify the simplification of GLAD by evaluating the anomaly detection performance of the GLAD$^0$ model. We adopt similar experiment set-up for GLAD$^0$ in order to test whether GLAD$^0$ can successfully detect the injected group anomalies.  As shown in Figure \ref{fig:glad0_robustness}, for most of the cases (expect for group number 8 and 9), GLAD$^0$ achieves the highest detection accuracy, while the other two-stage approaches are relatively unstable. Given the complexity of the model and the limited observations we feed in, the gain from GLAD$^0$ is less than that from the GLAD model. The performance deterioration with respect to the number of groups is due to the sparsity of the data. As we increase the group number of a fix size network, each group has fewer number of people, thus learning the role mixture for the group becomes more difficult.

We also report the simulation results on group anomaly detection for d-GLAD. The data is generated according to Algorithm \ref{alg:genDGLAD} with 5 time stamps. We manipulate the mixture rate of $50\%$ of the groups at time point 4 as injected anomalies. Then we raise alarms if the group's mixture rate deviates from the previous time by a certain threshold. In Figure \ref{fig:dglad_fp}, we display the false positive rate with different threshold values. For comparison, we train MMSB-MGM and GLAD at each time independently as baselines. It can be seen that d-GLAD achieves the lowest false positive rate, which demonstrates the gain of d-GLAD over static models on the dynamic dataset.

\subsubsection{Benchmark Data with Anomaly Labels}
\begin{figure*}[htbp]
\centering
\subfigure[Precision]{
\includegraphics[width=52mm ]{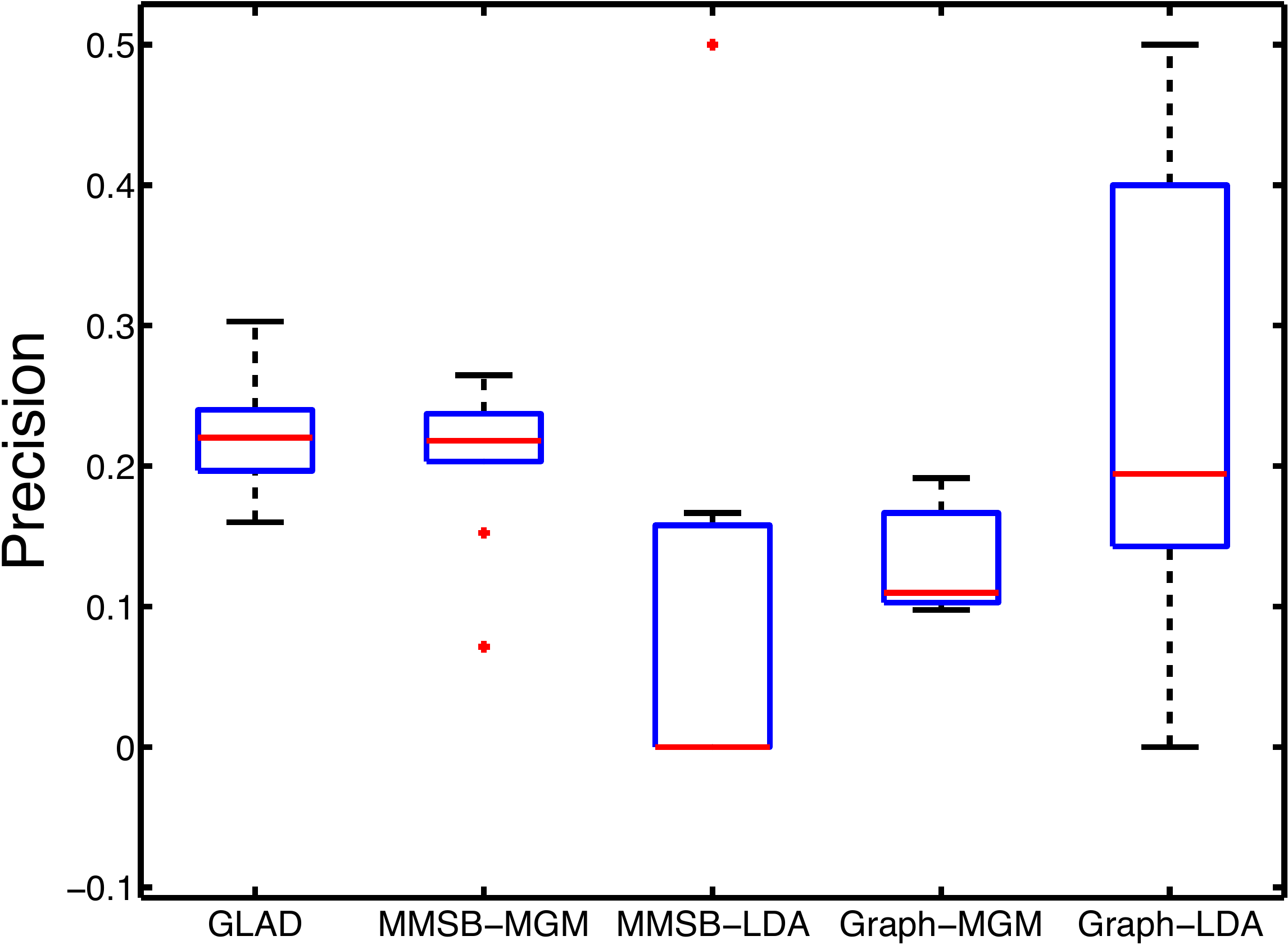} \label{fig:adams_precision}
}
\subfigure[Recall]{
\includegraphics[width=52mm ]{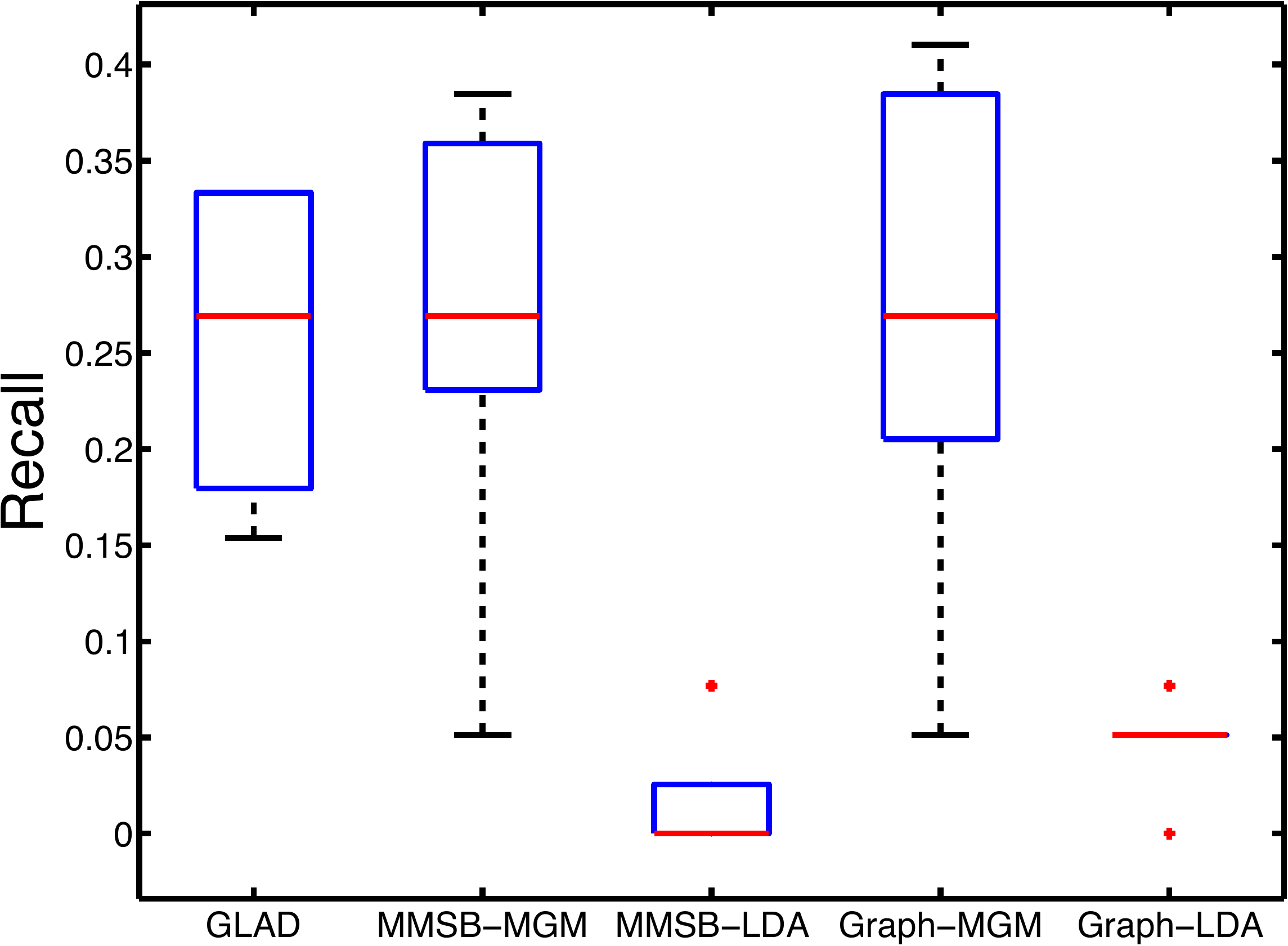} \label{fig:adams_recall}
}
\subfigure[F1 Score]{
\includegraphics[width=52mm]{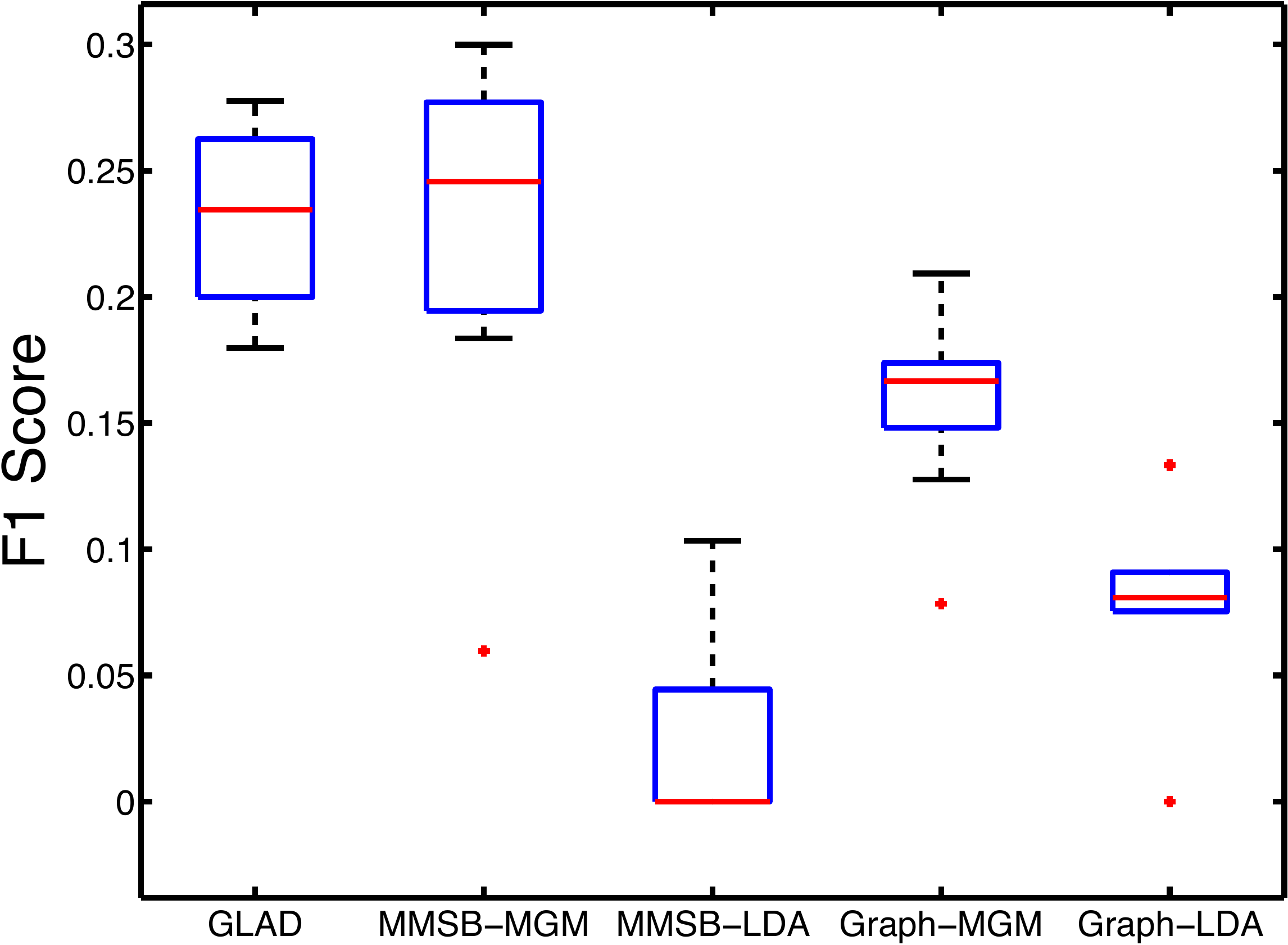}  \label{fig:adams_F1score}
}
\caption{\subref{fig:adams_precision}: Precision  \subref{fig:adams_recall}: Recall \subref{fig:adams_F1score}: F1 score on the benchmark dataset of GLAD and four baseline methods over 20 runs. All members in the anomalous groups are treated as individual anomalies and compared with 39 true anomalies. }
\label{fig:adams_detection}
\end{figure*}

The benchmark data set is generated by a simulator from a federal funded program. It contains email communication records and working activities from 258 company employees. Each employee is featured by 6 types of activities. The labeled dataset contains 39 individual anomalies and 5 of them cannot be detected by any existing algorithms. We set the number of groups as 20 as the optimal setting obtained from cross validation and calculate the anomaly score of each group by MCMC sampling. We treat all members in the most anomalous group as individual anomalies and compare them with the anomaly labels. Though the anomaly labels are point anomalies rather than group anomalies, the anomaly detection result reflects the potential of our approach to tackle other type of difficult anomaly detection problems. The precision, recall and F1 score over 20 runs on the benchmark dataset is shown in Figure \ref{fig:adams_detection}.

We can see that the GLAD model achieves comparable precision and recall with low variances. In contrast, the detection performances of the two-stage models fluctuates significantly. In terms of the F1 scores in Figure \ref{fig:adams_F1score}, both GLAD and MMSB-MGM beat the other algorithms while GLAD has a lower variance than MMSB-MGM. One possible explanation is that the point-wise features prevent the size of the group to become either too large or too small, thus leading to more robust performance.

\begin{table*}[ht]
\centering
\caption{Group Anomaly Accuracy of GLAD and four baselines on DBLP publications. With KDD papers treated as normal groups and other conferences are treated as group anomalies respectively.}
\label{tb:dblp_anomaly}
\begin{tabular}{|c|c|c|c|c|c|} \hline
Methods & GLAD & Graph-LDA & Graph-MGM & MMSB-LDA & MMSB-MGM  \\ \hline
\texttt{DBLP:KDD/CVPR} & \textbf{0.4167} & 0.3333 & 0.3333 &  0.2500 & 0.2500  \\ \hline
\texttt{DBLP:KDD/ICML} & \textbf{0.2500} & 0.0833 & 0.0833 & 0.1667  & 0.1667  \\ \hline
\texttt{DBLP:KDD/SIGMOD} & \textbf{0.2875} & 0.0750 & 0.0500 & 0.1625  & 0.1625  \\ \hline
\texttt{DBLP:KDD/CIKM} & \textbf{0.4500} & 0.4000 & 0.3625 & 0.2625  & 0.2625  \\ \hline
\texttt{DBLP:KDD/EDBT} & \textbf{0.2625} & 0.0500 & 0.0875 & 0.2000 & 0.2000 \\ \hline
\end{tabular}
\end{table*}

\subsection{Real World Datasets}
\subsubsection{Scientific Publications}
Researchers study the topics of papers seeking for concise representations of scientific publications, which contain both pair-wise data like co-authorship and point-wise data such as bag of words features. Detecting anomalous topic distributions in scientific publications can sharpen our understanding of the structure of research communities and possibly reveal unusual research trends.  In order to quantify our method, we resort to anomaly injection and construct a dataset with group anomaly labels. One way to construct group anomalies is the scenario that a conference paper corpus is contaminated by group of papers from conferences in other domains. 

We create a dataset from a pre-processed Digital Bibliography and Library Project (DBLP) dataset from \cite{deng2011probabilistic}. The dataset consists of conference papers from 20 conferences of four major area: database (DB), data mining (DM), information retrieval (IR) and artificial intelligence (AI). Each paper has a bag-of words feature vector with a vocabulary size of 11,771 and associated 28,702 authors information. The detailed statistics of the dataset are shown in the top half of Table \ref{tb:dblp_statistics}. We set up the group anomaly detection scenario as follows: we randomly sample groups of papers from KDD and treat them as normal groups. Then we sample groups of papers from the other conferences (e.g, CVPR, ICML , SIGMOD) and inject them into KDD papers as group anomalies. If the two papers have at least one common author, we add a link between them.

Accordingly, all conferences share four topics. But different conferences might have difference point of emphasis, resulting in different mixture rates of topics. Our goal is to pick out the ``anomalous'' papers from the corpus. We sample 50 groups of papers and inject $20\%$ group anomalies. We apply different models with 50 groups and 4 roles to the data for inference of the membership and role distributions. Then we rank 50 groups with respect to their anomaly scores. We treat the top $20\%$ groups as the detected anomalies. Table \ref{tb:dblp_anomaly} shows the anomaly detection accuracy by GLAD and four other baselines. GLAD is superior to all four baselines models for  different combination of normal/abnormal settings. We also display the topics learned by the GLAD model. In Table \ref{tb:dblp_topics}, we show the top ten most representative words for the four topics, which well reproduce the topic results reported in \cite{deng2011probabilistic}.

\begin{table}[h]
\centering
\caption{Key statistics of the DBLP and ACM publication datasets}
\label{tb:dblp_statistics}
\begin{tabular}{|c|c|c|c|} \hline
\multicolumn{4}{|c|}{DBLP} \\ \hline
\# \texttt{of docs} & 28,569 &  \# \texttt{of authors} & 28,702 \\ \hline
\# \texttt{ of conf} & 20 &  \# \texttt{of words} & 11,771 \\ \hline
 \# \texttt{of links}  &  104,962 & \# \texttt{  of area } & 4 \\ \hline
\multicolumn{4}{|c|}{ACM} \\ \hline
 \#  \texttt{  of docs }  & 31,574 &  \# \texttt{ of authors }& 4,474\\ \hline
 \# \texttt{ of year } & 10  &  \# \texttt{ of words} & 8,024\\ \hline
\end{tabular}
\end{table}

\begin{table}[h]
\centering
\caption{The most representative words learned by GLAD on DBLP dataset of four topics: database, data mining, information retrieval and artificial intelligence.}
\label{tb:dblp_topics}
\begin{tabular}{|c|c|c|c|} \hline
\texttt{DB} & \texttt{DM} & \texttt{IR} & \texttt{AI} \\ \hline
databases & data   & web   &  query\\
object &  mining   & information &  system\\
access &  efficient & learning & management  \\
database & query & search & processing \\
oriented & algorithm & retrieval & web \\
security & queries & clustering & efficient \\
based & clustering & query & performance \\
indexing & databases & text & infomation \\
systems & algorithms & model & distributed \\
privacy & large & classification & optimization \\
\hline
\end{tabular}
\end{table}

Since the DBLP dataset does not contain time-specific information which is not suitable for the d-GLAD model, we process another ACM dataset downloaded from ArnetMiner \cite{Tang:08KDD}. The dataset contains the publications from year 2000 to 2009 by 4,474 authors, mainly from the data mining community. In order to study the topic evolution for academic scholars, we extract the abstracts of all publications and group them by authors and publishing years. For each author, we construct a bag of words feature vector out of all the papers he/she has written in one year. And the communication networks we generate are based on the co-authorship of the papers. Whenever two authors have collaborations in a certain year, we create a link between them for the network snapshot in that year.

Due to the lack of labels, it is difficult to directly evaluate our model on anomaly detection task. As an alternative, we design a prediction task to compare the modeling performance of GLAD and d-GLAD on ACM publications. Specifically, we separate the papers into training and testing sets and measure the predictive model log-likelihood on the testing data. For d-GLAD, we train our model using a series of publications from previous years, and test on the year immediately after. For the GLAD model, as it is a static model, time independence assumption applies.  We train the model using previous year and test on the next year. The model fitting results are shown in Table \ref{tb:prediction_likelihood}. Out of 9 training-testing experiments, d-GLAD model achieves higher log-likelihood than GLAD model for 6 times, indicating d-GLAD as a better fit for the evolving publication modeling.
\begin{table*}
{\small
\centering
\caption{Prediction negative log likelihood for GLAD and d-GLAD on ACM dataset over 9 years. }
\begin{tabular}{|c|c|c|c|c|c|c|c|c|c|}
\hline
\texttt{Year} & 2001 & 2002 &  2003 & 2004 & 2005 & 2006 & 2007 & 2008 & 2009 \\ \hline
\texttt{GLAD } &28421.63 &	28023.68  & \textbf{30184.66} &	\textbf{32039.92} &28317.67	& \textbf{30539.66}
 & 26105.21 &	34340.53 & 25967.75\\ \hline
\texttt{DGLAD } & \textbf{34411.28 }& \textbf{33411.14} & 29935.87 	 &	31958.92 &\textbf{30082.65}	& 29696.12 & \textbf{30042.77} &\textbf{34395.68} & \textbf{31683.49} \\ \hline
\end{tabular}
\label{tb:prediction_likelihood}
}
\end{table*}

\subsubsection{US Senate Voting}
We collect the voting records from the government website of United States 109th Congress \footnote{http://www.senate.gov/} using the New York Time Congress API  \footnote{http://developer.nytimes.com/docs/read/congress\_api}. The records of 109th Congress contain 100 senators' voting spanning two sessions from Jan 1st 2005 to Dec 31st 2006. We divide the 24 months records into 8 time slots, where each slot denotes a 3-month interval. Then we apply the method of \cite{kolar_estimating_2008} to construct a network from original yay/nay votes. For the nodes features, we collect the statistics of  votes in six dimensions, namely House Joint Resolution(hjres), House of Representatives(hr), Presidential Nomination(pn), Simple Resolution(s), sconres(Senate Concurrent Resolution) and Senate Joint Resolution(sjres). We evaluate GLAD on single aggregated network and d-GLAD on the 8 time slots time-varying data.

\begin{figure}[t!]
\centering
\includegraphics[scale = 0.5]{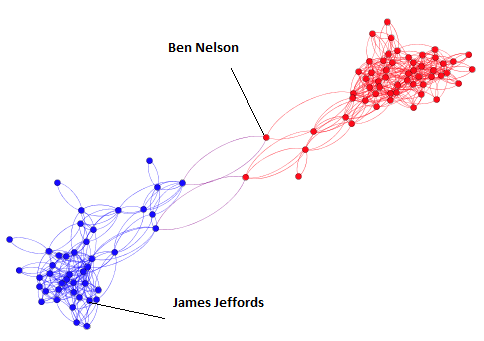}
\caption{ Common votes graph with party labels inferred by GLAD for 100 senators on the aggregated network. Compared with ground truth, two outliers are highlighted due to their anomalous voting behavior.}
\label{fig:senatorGraph}
\end{figure}

We set the number of groups as 2 and number of roles as 3 as  the Senate consists of two major parties and maintains three types of committees. Figure \ref{fig:senatorGraph} shows the groups inferred by GLAD. The blue nodes denote Democratic party members and the red ones are Republican. Compared with known facts, the model correctly reveals the party affiliation except for two outliers: Ben Nelson (Democratic) and James Jeffords (Independent). The underlying reason is that the votes of these two senators are often at odds with the leadership of his party, leading to false grouping.  We conduct an anecdotal investigation and find that the congressional vote rating from the National Journal placed Ben Nelson to the right of five Senate Republicans in 2006. For James Jeffords, he served as a Republican until 2001, when he left the party to become an Independent and began caucusing with the Democrats.

Since there are merely two groups, it is impetuous to say one party is more anomalous than the other. Instead, we use d-GLAD to detect time points when the role mixture rates change dramatically. In fact, d-GLAD raises an alarm at the 7th time-step for Democratic. A well known political event happened during this time is that Democratic senator Joseph Lieberman lost the Democratic Party primary election and became a independent Democratic in September 2006. Though it may be over-optimistic to draw the conclusion that this event causes the sudden change of role mixture rates, it serves as an evidence that the dynamics of the voting behavior is closely related to the party affiliation of members.

%\begin{figure}[t!]
%\centering
%\includegraphics[scale = 0.23]{./party.png}
%\caption{Party affiliation inferred by GLAD for 50 sample senators on the dynamic network over 8 time stamps. The top letter [R/D] is the ground truth label.}
%\label{fig:senatorGraph}
%\end{figure}

%\section{Discussion}
%\label{sec:discussion}
%\input{discussion.tex}

\section{Conclusion}
\label{sec:conclusion}
In this paper, we perform a follow-up study of the Group Latent Anomaly Detection (GLAD) model by analyzing an alternative construction of the unified model. We loosely connect the MMSB model and the LDA model assuming the shared group membership distribution for both point-wise and pair-wise data. We also provide the  variational Bayesian inference algorithm for model inference. We conduct a simulation experiment to verify  the benefit of the joint model in comparison with the two-stage approaches. 

\section{Acknowledgments}
\label{sec:acknowledge}
The research was sponsored by the U.S. Defense Advanced Research Projects Agency (DARPA) under the Anomaly Detection at Multiple Scales (ADAMS)  program, Agreement Number W911NF-11-C-0200 and NSF research grants IIS-1134990. The views and conclusions are those of the authors and should not be interpreted as representing the official policies of the funding agency, or the U.S. Government.

%\appendix
%\section{variational Bayes inference of GLAD$^0$}
%\label{sec:inferGLAD0}
%\input{inference}

\bibliography{arxiv.bib}
\bibliographystyle{abbrv}

\end{document}